\documentclass{article}

\usepackage{microtype}
\usepackage{graphicx}
\usepackage{subfigure}
\usepackage{booktabs} 
\usepackage{amsmath}
\usepackage{amssymb}
\usepackage{enumitem}
\usepackage{multirow}
\usepackage{units}
\usepackage{appendix}

\usepackage{hyperref}

\usepackage{amsmath}
\usepackage{amssymb}
\usepackage[accepted]{icml2020}

\icmltitlerunning{On Class Orderings for Incremental Learning}

\newcommand{\figaccuracies}{
\begin{figure}[t]
\begin{center}
  \includegraphics[width=0.95\linewidth]{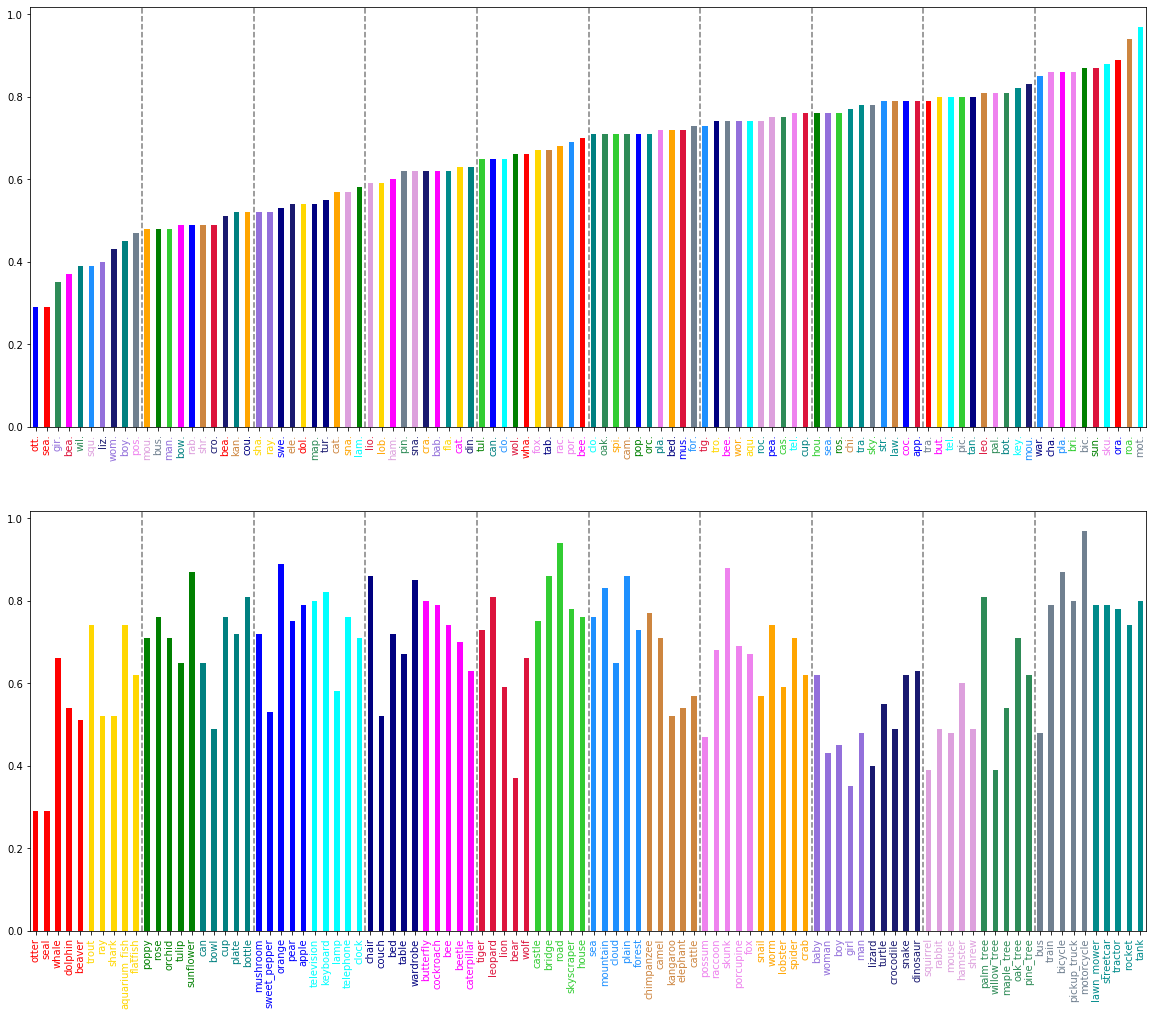}
  \vspace{-0.8em}
  \caption{Accuracies of CIFAR-100 classes after a single non-incremental training on ResNet-32 incrementally ordered by class accuracy (top), and grouped with provided coarse-grained labels (bottom). Dashed lines represent task boundaries for an equally divided 10-task split. Colors denote coarse-grained group labels.}
  \label{fig:per_class_accs}
  \vspace{-0.8em}
\end{center}
\end{figure}
}
\newcommand{\figconfmat}{
\begin{figure}[t]
\begin{center}
  \subfigure{
  {\includegraphics[width=0.49\linewidth]{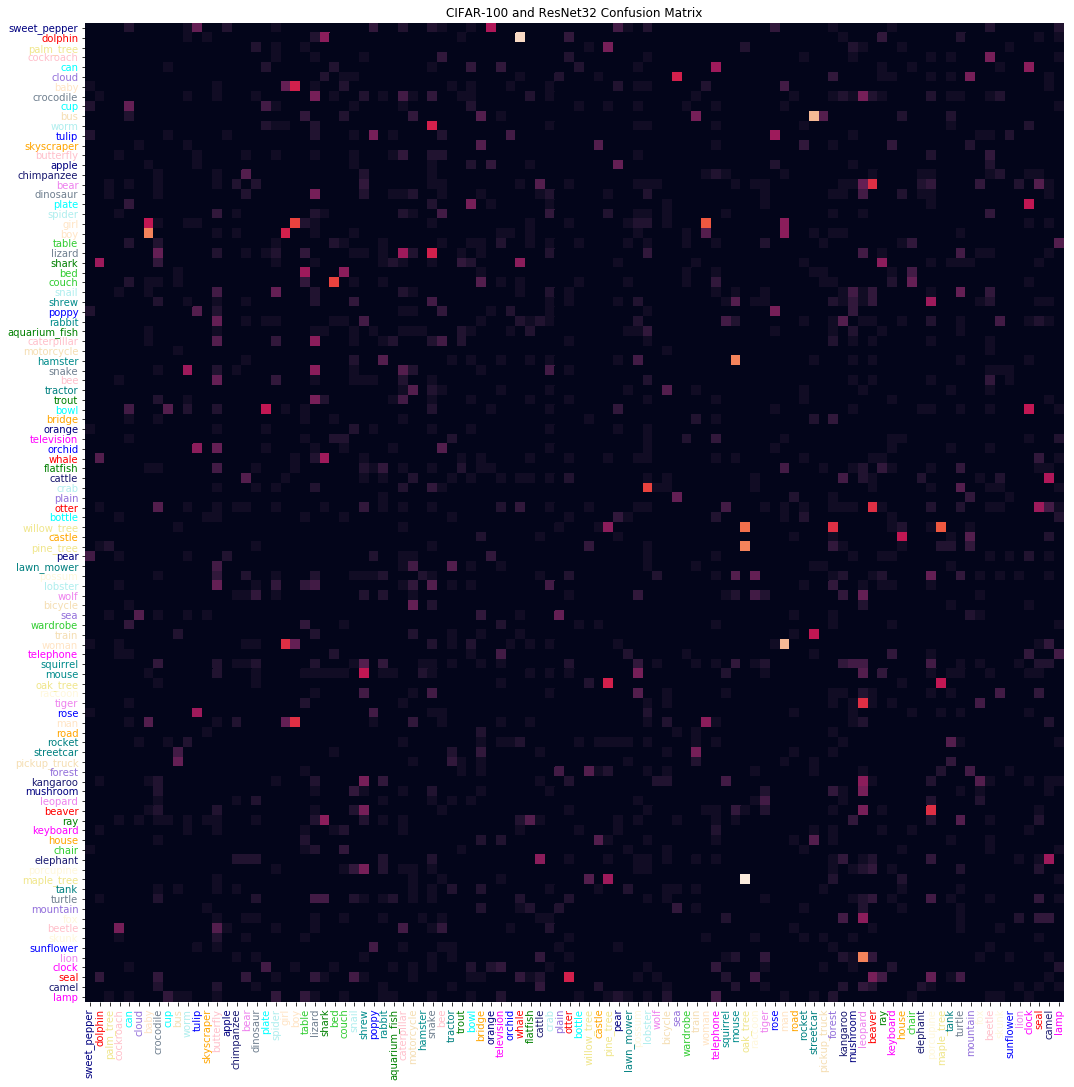}}%
  }\,
  \subfigure{
  {\includegraphics[width=0.49\linewidth]{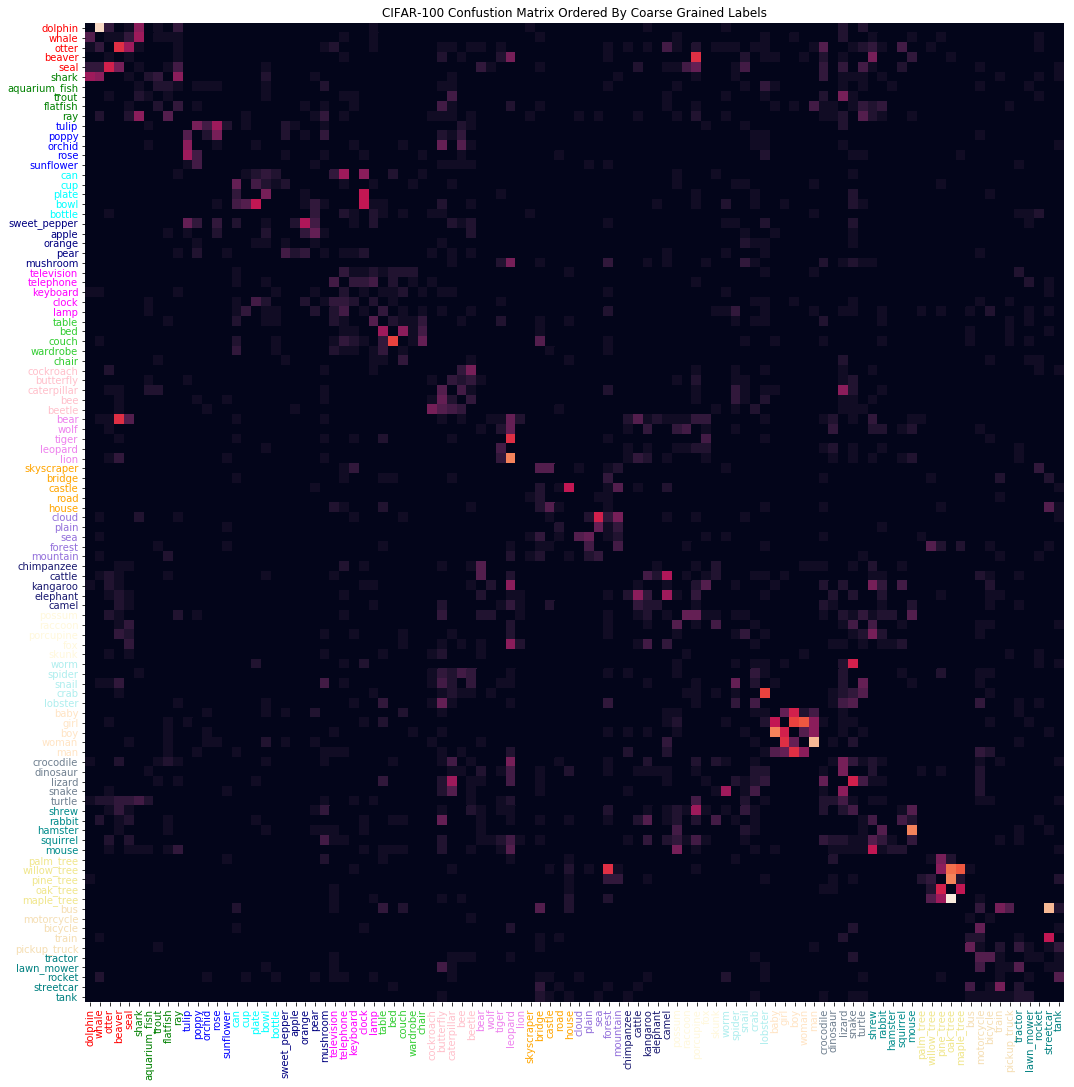}}%
  }
  \vspace{-1.5em}
  \caption{CM from CIFAR-100 with original class order (left) and coarse grained labels order (right) after a joint training on \mbox{ResNet-32}. Diagonal values are skipped for better visualization.}
  \label{fig:conf_mat}
  \vspace{-1.1em}
\end{center}
\end{figure}
}
\newcommand{\figclassorders}{
\begin{figure}[t]
\begin{center}
  \subfigure{%
  {\centering\includegraphics[trim=7 7 6 6, clip, width=0.32\linewidth]{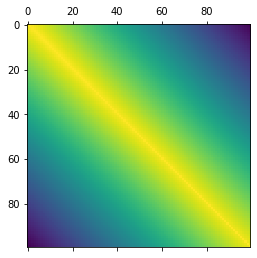}}%
  }\,
  \subfigure{%
  {\centering\includegraphics[trim=7 7 6 6, clip, width=0.32\linewidth]{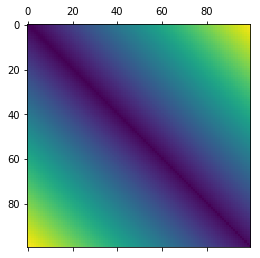}}%
  }\,
  \subfigure{%
  {\centering\includegraphics[trim=7 7 6 6, clip, width=0.32\linewidth]{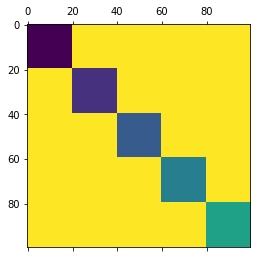}}%
  }\\
  \vspace{-0.4em}
  \subfigure{%
  {\centering\includegraphics[trim=7 7 6 6, clip, width=0.32\linewidth]{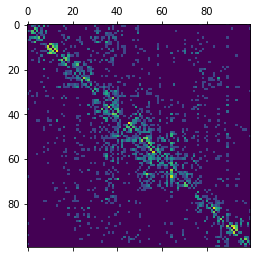}}%
  }\,
  \subfigure{%
  {\centering\includegraphics[trim=7 7 6 6, clip, width=0.32\linewidth]{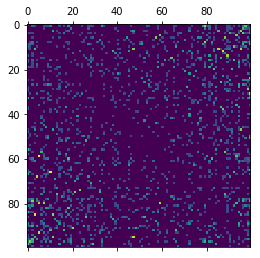}}%
  }\,
  \subfigure{%
  {\centering\includegraphics[trim=7 7 6 6, clip, width=0.32\linewidth]{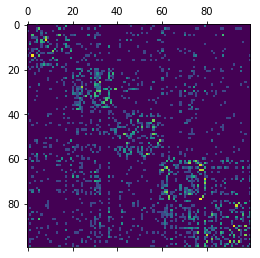}}%
  }
  \vspace{-1.2em}
  \caption{Class ordering objectives (top) and resulting confusion matrices (bottom).}
  \label{fig:class_order}
\end{center}
\vspace{-0.8em}
\end{figure}
}

\newcommand{\figcifarnone}{
\begin{figure*}[t]
\begin{center}
  \subfigure{%
  {\centering\includegraphics[width=0.49\linewidth]{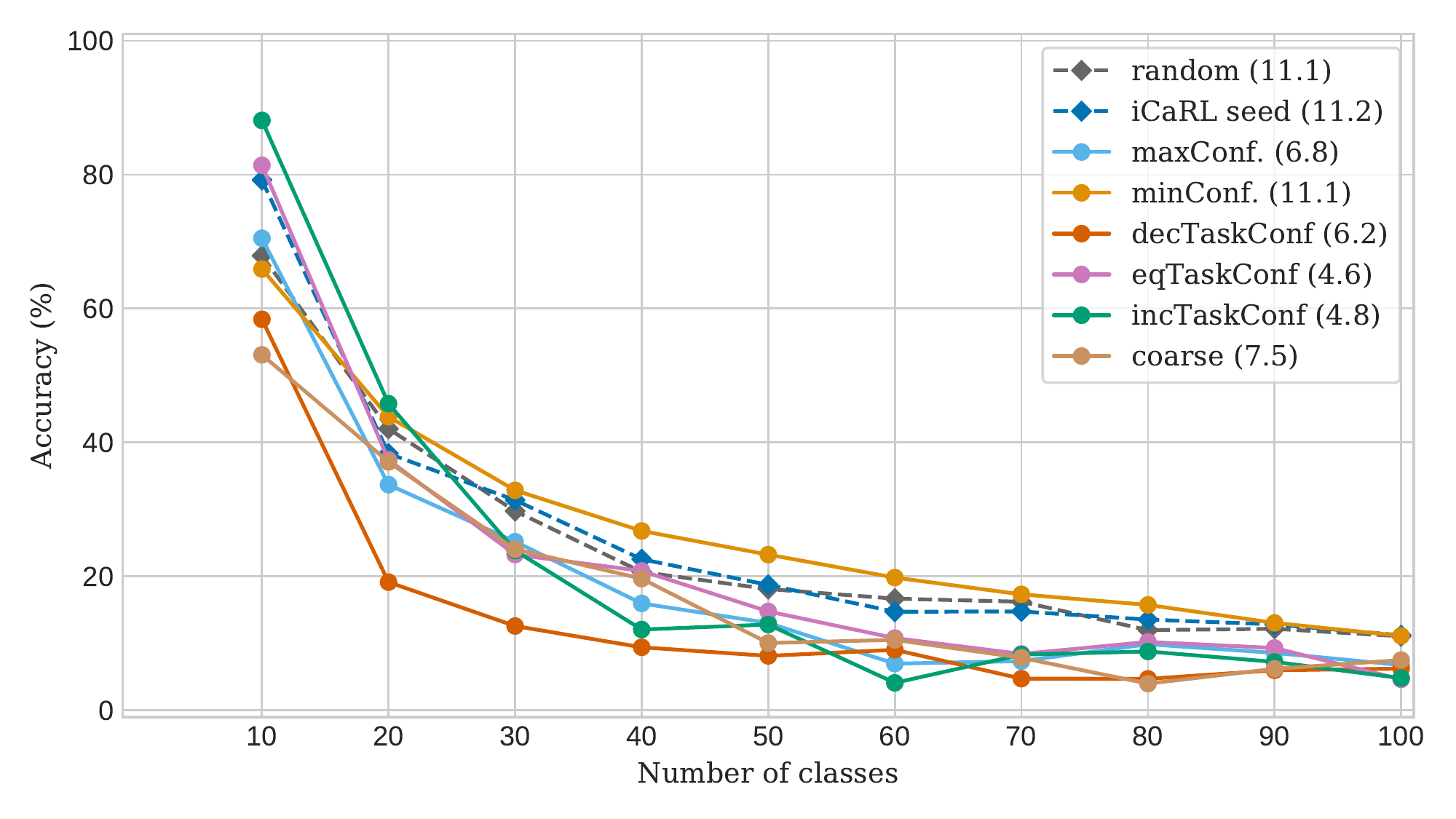}}%
  }\,
  \subfigure{%
  {\centering\includegraphics[width=0.49\linewidth]{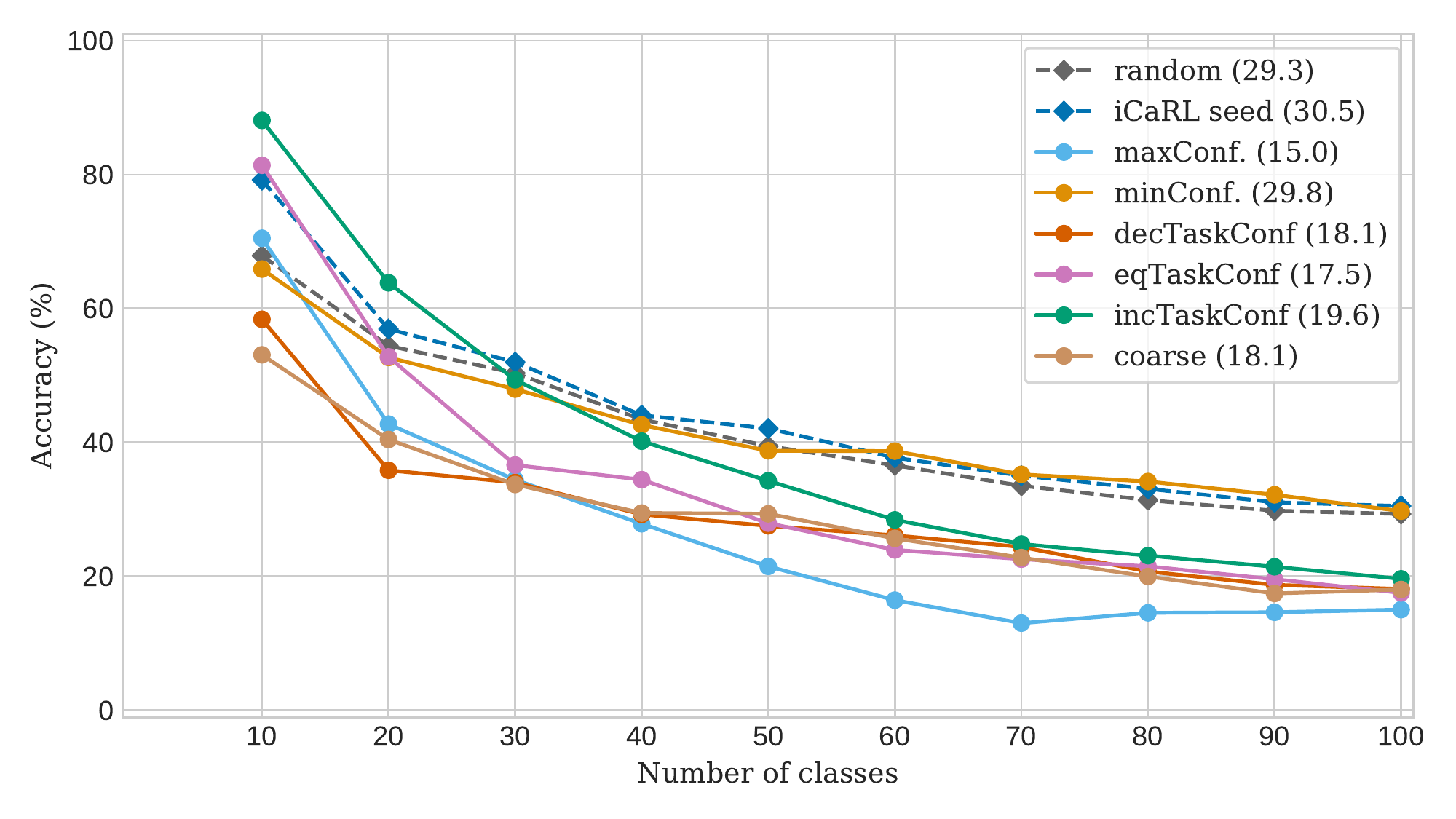}}%
  }
  \vspace{-1.0em}
  \caption{FT (left) and LwF (right) with different class orderings for CIFAR-100 on ResNet-32 from scratch without exemplars memory.}
  \label{fig:cifar_none}
  \vspace{-0.8em}
\end{center}
\end{figure*}
}
\newcommand{\figExpSingleColumn}[3]{
\begin{figure}[h!]
\begin{center}
  \includegraphics[trim=11 11 12 12, clip, width=#1\linewidth]{./figures/#2.pdf}
  \vspace{-1.5em}
  \caption{#3}
  \label{fig:#2}
\end{center}
\end{figure}
}

\begin{document}
\twocolumn[
\icmltitle{On Class Orderings for Incremental Learning}
\icmlsetsymbol{equal}{*}
\begin{icmlauthorlist}
\icmlauthor{Marc Masana}{equal,cvc}
\icmlauthor{Bart{\l}omiej Twardowski}{equal,cvc}
\icmlauthor{Joost van de Weijer}{cvc}
\end{icmlauthorlist}
\icmlaffiliation{cvc}{LAMP team, Computer Vision Center, UAB Barcelona, Spain}
\icmlcorrespondingauthor{Marc Masana}{mmasana@cvc.uab.cat}
\icmlkeywords{Machine Learning, ICML}
\vskip 0.3in
]

\printAffiliationsAndNotice{\icmlEqualContribution}

\begin{abstract}
The influence of class orderings in the evaluation of incremental learning has received very little attention. In this paper, we investigate the impact of class orderings for incrementally learned classifiers. We propose a method to compute various orderings for a dataset. The orderings are derived by simulated annealing optimization from the confusion matrix and reflect different incremental learning scenarios, including maximally and minimally confusing tasks. We evaluate a wide range of state-of-the-art incremental learning methods on the proposed orderings. Results show that orderings can have a significant impact on performance and the ranking of the methods.
\end{abstract}

\section{Introduction}
\label{sec:introduction}

Incremental learning (IL) has gained popularity over the last years as a way to continuously introduce new concepts to an existing model. Incrementally learning tasks relieves the issues of maintaining and retraining large datasets, and costs associated with it. However, retaining knowledge when retraining on different data in artificial neural networks is not a trivial task. Whenever a network is trained only on new data, the abrupt loss of previously acquired knowledge manifests. This phenomenon is know as \emph{catastrophic forgetting}~\cite{mccloskey1989catastrophic}. In recent years, many methods have been proposed to alleviate this problem which stands in the way to advanced life-long learning systems~\cite{lesort2020continual, de2019continual, parisi2019continual}.

The problem of continual learning is often simplified to an incremental learning of new concepts (classes) in a well-defined, equally divided, sequence of tasks. This may sound artificial, but is a common choice in recent works~\cite{aljundi2016expert, li2017learning, rebuffi2017icarl, chaudhry2018riemannian, belouadah2019il2m}. Having the same experimental settings makes it easy to compare results and performance. Specifically, class-incremental learning (class-IL) defines the setting where the task-ID is unknown at inference time, making the predictions of the learned models task-agnostic. Most class-IL methods consider multi-class incremental learning~\cite{hou2019learning}, where only tasks that at least have two or more classes are considered. Those also conform to the prevalent practise of making all tasks have the same number of classes. This rises the question about data preparation: are we missing something important when ordering the tasks and classes? In this work we investigate how relevant is class ordering and how it can affect the results of well-known class-IL methods.

\figaccuracies

It has become common to compare different approaches by using CIFAR-100, while recently some other larger datasets have been getting more exposure. In Fig.~\ref{fig:per_class_accs}, we show that class ordering can influence task accuracy within a particular split for an already trained model. Therefore, some influence can be expected to multi-class-IL. Overall performance---measured as a value of mean avg. accuracy across all tasks after reaching the last one---can depend on how hard is the ordering. In our research, we propose a method based on confusion matrix (CM) ordering (see Fig.~\ref{fig:conf_mat}) that helps explore the \emph{difficulty} of the incrementally learned classifier even further.

The main contributions of this paper are: 1) proposing a novel method for class ordering in IL scenarios based on confusion matrix values, 2) investigating IL methods robustness to class ordering, 3) analysing of some commonly used split strategies in comparison to the random ones.

\figconfmat

\section{Related work}
\label{sec:relatedwork}

\textbf{Class-IL:} We chose some class-IL methods that are common comparison in the literature, and some that are current state-of-the-art. LwF~\cite{li2017learning} is a regularization-based method which adds a constrain loss to the outputs of older classes to not change too much when learning a new task. Similarly, EWC~\cite{kirkpatrick2017overcoming} also applies a regularization constrain on the weights to limit their shift. iCaRL~\cite{rebuffi2017icarl} proposed to extend LwF by keeping a small memory with data exemplars which is replayed during training. Following this idea, BiC~\cite{wu2019large} and LUCIR~\cite{hou2019learning} extend the usage of distillation and exemplars to also apply a bias correction to the outputs of different tasks, allowing to compensate the imbalance introduced by new tasks. Finally, IL2M~\cite{belouadah2019il2m} also proposes bias correction, although over Finetuning since they show that it works better than applying it over LwF.

\textbf{Class ordering:} Most works on class-IL report results by using a random order of classes on CIFAR-100, i.e. iCaRL, LUCIR, BiC. Furthermore, the popularity of iCaRL and the interest in comparing with it, makes quite common their specific class ordering (their code fixes the random seed to 1993). However, none of them look deeper into the choosing of that specific class order. In~\cite{masana2020ternary} and~\cite{de2019continual}, the authors touch the subject of class ordering, showing that some methods report different results based on different class orderings. Only a random ordering and a semantically split ordering were investigated, without any dedicated method to order classes harder or easier, as this was not their main focus.

\textbf{Curriculum Learning and Classification Complexity:}
In contrast to curriculum learning, our objective is not to obtain the best model performance~\cite{Pentina2015curriculum}, but to propose an evaluation for class-IL methods under different scenarios. Another similar research direction is assessing how complex is a classification task. In our case, we could use a known measure~\cite{Lorena2019classcomplexity}, or the one proposed in~\cite{nguyen2019understanding} for IL.

\section{Class ordering}
\label{sec:classordering}

In multi class-IL for image classification, a sequence of tasks where each consists of $m_t$ classes is learned one at a time, extending the knowledge of the model in incremental steps. Given a set of paired data ${\bf x}_{i}$ with their respective class labels $y_{i} \in C^{t}$, where $C^t=\{c^t_1,c^t_2,\ldots,c^t_{m^t}\}$ denotes the set of $m^{t}$ classes of task $t$. When training on task $t$, only data $\left({\bf x}_{i}, y_{i}\right)\sim D^t$ is available. We consider disjoint classes between all tasks, $C^t \cap C^s=\varnothing $ for $t \neq s$ as in~\cite{aljundi2016expert, chaudhry2018riemannian, dhar2019learning, hou2019learning, liu2018rotate, rebuffi2017icarl,  yu2020semantic}. After training each task we evaluate the learned model on all classes seen so far $C=\bigcap_{i < t} C^t$.

Let $M\in\mathbb{N}^{|C|\times|C|}$ be the confusion matrix from learning all classes in a non-incremental way (usually known as \emph{joint training}), where each element $M_{ij}$ defines how many times samples of class $i$ are predicted as class $j$. Based on the information contained in $M$, we can estimate how \emph{confusing} class $i$ is by looking at how much it gets wrongly classified as any another class $j$. Consequently, also how easy or difficult they are to tell apart when having all data available during the training session. We assume that any advantage that comes by having all information available can be mitigated or removed when changing to an IL setting. In joint training, specific features in the network can be learned to focus on differentiating two classes that can easily be confused. However, in an IL setting, those discriminative features become more difficult to learn or can be modified afterwards, especially when the classes belong to different tasks. Thus, the difficulty of the task can be perceived differently in each scenario. Depending on the method, it may handle this issue differently, and therefore lead to more catastrophic forgetting.

\figclassorders

As a result, we define different class orderings. In order to illustrate the effect of class ordering, we take the CIFAR-100 dataset~\cite{krizhevsky2009learning} as an example, although the proposed strategies are extensible to other datasets. First, we define the baseline class ordering as:

\vspace{-0.8em}
\begin{itemize}
    \item \textbf{random}: takes a permutation of the class order. By default taking the original class ordering which the dataset provides (usually alphabetically ordered or similar), or otherwise a permutation corresponding to a random seed. As explained in Sec.~\ref{sec:relatedwork}, in the case of CIFAR-100, some works decide to fix the seed to the same as iCaRL~\cite{rebuffi2017icarl}.
\end{itemize}
\vspace{-0.8em}
If we train a model in a single training session (joint training) with all data for all classes, we can calculate CM, as seen on Fig.~\ref{fig:conf_mat}. Based on that, we define two more class orders as:

\vspace{-0.8em}
\begin{itemize}
    \item \textbf{max confusion} (maxConf): highly miss-classified classes are next to each other---max confusion is happening around the CM diagonal. This creates an IL split with more difficult intra-task classification.
    \item \textbf{min confusion} (minConf): enforces classes which are rarely miss-classified to be in the same task---max confusion is happening at the corners of the CM. Intra-task and adjacency tasks classification becomes easier, but also pushes the most miss-classified classes towards the first and last tasks.
\end{itemize}
\vspace{-0.8em}

Finding the above orderings based on the CM values is a non-trivial task, where a brute-force naive approach of $O(n!)$ cannot be applied even to moderate size problems. To reduce the complexity, we  re-formulate finding the class ordering as an optimization problem where the objective is to maximize the value of the fitness function to a desired $M$. Then, we use an objective weight matrix $W\in\mathbb{N}^{|C|\times|C|}$ (see Fig.~\ref{fig:class_order}) which is used to calculate a score value:
\begin{equation}
\text{score}(M, W) = \mathrm{tr}(W^T M),
\end{equation}
to assess adaptation in a search for a solution. In this case, a global optimum is not necessary since it is hard to establish. Instead, we use the simulated annealing optimization algorithm to find an ordering in a constrained time limited by the number of fitting iterations. The same approach has been previously used for CM ordering for better visualization of large matrices in~\cite{thoma2017analysis} and for HASYv2 dataset~\cite{thoma2017hasyv2}.

We can also define an objective CM, and try to converge to a permutation that better accommodates to incremental tasks by introducing their boundaries. If the splits of $C^{1},\ldots,C^{t}$ can be know upfront, which is usually the case, we can incorporate task boundaries in the weighting matrix for both scenarios. Therefore, we can define three more orderings:
\vspace{-0.8em}
\begin{itemize}
    \item \textbf{increasing task confusion} (incTaskConf): maximize confusion in all tasks around the diagonal of $M$ with increasing confusion between them. The objective matrix is presented in Fig.~\ref{fig:class_order} (right).
    \item \textbf{equal task confusion} (eqTaskConf): similar to max confusion, but introducing task boundaries should cause less confusion between adjacent tasks.
    \item \textbf{decreasing task confusion} (decTaskConf): maximize confusion in all tasks around the diagonal while decreasing it between them. Similar to increasing, but with inverted diagonal weights from Fig.~\ref{fig:class_order} (right).
\end{itemize}
\vspace{-0.8em}
In the specific case of CIFAR-100, since a coarse grained hierarchy of the classes exists, we can also define an ordering based on the provided two level taxonomy. In each task we can have classes related to the same group or similar groups in order to make the classification harder.
\vspace{-0.8em}
\begin{itemize}
    \item \textbf{coarse grained}: ordered by a provided grouping or taxonomy. For CIFAR-100 classes are divided into 20 groups, as shown with bar colors in Fig.~\ref{fig:per_class_accs} and labels in Fig.~\ref{fig:conf_mat} (right).
\end{itemize}
\vspace{-0.8em}

\begin{table*}[tb]
\caption{CIFAR-100 results for class-IL with growing memory of 20 exemplars per class (10 runs average and standard deviation). The best score for each task of each method is in bold. Underscore marks the lowest score.}
\label{tab:exemplars-growing}
\begin{center}
\resizebox{0.99\linewidth}{!}{
\begin{tabular}{lccccccccc}
\toprule
        &  task  &                    Random &             iCaRL seed &                    coarse &                maxConf &                   minConf &               decTaskConf &             eqTaskConf &            incTaskConf \\
\midrule
LwF & 2 &              51.8$\pm$5.3 &           50.8$\pm$2.6 &              45.9$\pm$2.7 &           52.4$\pm$2.0 &              44.6$\pm$1.9 &  \underline{41.5}$\pm$3.2 &           53.1$\pm$3.9 &  \textbf{62.5}$\pm$2.7 \\
        & 5 &  \underline{31.8}$\pm$3.9 &           33.2$\pm$3.3 &              36.3$\pm$2.2 &           33.0$\pm$2.9 &              33.8$\pm$2.3 &              33.9$\pm$2.6 &           35.9$\pm$3.2 &  \textbf{37.4}$\pm$2.9 \\
        & 10 &  \underline{24.8}$\pm$2.6 &           27.3$\pm$2.2 &              26.5$\pm$2.0 &  \textbf{32.6}$\pm$3.1 &              25.4$\pm$1.4 &              29.9$\pm$2.4 &           31.6$\pm$3.3 &           29.4$\pm$1.9 \\
\midrule
iCaRL & 2 &              61.4$\pm$3.8 &           58.7$\pm$3.6 &              49.9$\pm$3.1 &           55.1$\pm$1.6 &              55.5$\pm$1.3 &  \underline{44.9}$\pm$2.4 &           60.0$\pm$2.7 &  \textbf{67.2}$\pm$3.9 \\
        & 5 &              42.0$\pm$3.6 &           42.5$\pm$2.7 &              41.7$\pm$2.2 &           39.6$\pm$2.3 &              43.7$\pm$2.4 &  \underline{33.4}$\pm$2.6 &           44.5$\pm$2.4 &  \textbf{45.9}$\pm$3.6 \\
        & 10 &              33.8$\pm$3.7 &           34.2$\pm$2.6 &              32.8$\pm$1.7 &  \textbf{35.4}$\pm$2.6 &              33.4$\pm$2.3 &  \underline{28.0}$\pm$2.2 &           35.4$\pm$3.0 &           32.0$\pm$3.1 \\
\midrule
BiC & 2 &              61.2$\pm$5.3 &           56.4$\pm$3.8 &              50.4$\pm$3.0 &           52.8$\pm$2.6 &              51.5$\pm$2.7 &  \underline{41.2}$\pm$2.1 &           57.9$\pm$4.1 &  \textbf{69.7}$\pm$2.7 \\
        & 5 &              44.8$\pm$3.2 &           45.2$\pm$3.0 &              44.4$\pm$2.6 &           40.7$\pm$3.7 &              47.7$\pm$1.3 &  \underline{37.1}$\pm$2.9 &           45.3$\pm$3.1 &  \textbf{52.4}$\pm$2.8 \\
        & 10 &              39.3$\pm$1.9 &  40.1$\pm$2.8 &              39.3$\pm$2.3 &           37.5$\pm$2.7 &              \textbf{40.1}$\pm$1.3 &  \underline{37.2}$\pm$2.9 &           38.6$\pm$2.0 &           37.8$\pm$2.6 \\
\midrule
LUCIR & 2 &              63.2$\pm$3.4 &           59.6$\pm$3.3 &              53.4$\pm$2.9 &           54.0$\pm$3.2 &              53.3$\pm$3.3 &  \underline{47.2}$\pm$1.9 &           61.2$\pm$2.0 &  \textbf{72.0}$\pm$2.2 \\
        & 5 &              40.0$\pm$3.4 &           40.7$\pm$3.5 &              40.3$\pm$2.6 &           37.2$\pm$5.6 &              39.5$\pm$2.8 &  \underline{36.5}$\pm$2.9 &           45.8$\pm$1.4 &  \textbf{47.3}$\pm$1.9 \\
        & 10 &              27.2$\pm$2.7 &           29.6$\pm$3.2 &  \underline{26.2}$\pm$3.1 &           28.9$\pm$5.8 &              27.7$\pm$1.9 &              29.1$\pm$3.3 &  \textbf{31.9}$\pm$2.0 &           28.5$\pm$1.6 \\
\midrule
IL2M & 2 &              58.2$\pm$5.1 &           52.2$\pm$3.7 &              43.2$\pm$5.9 &           51.2$\pm$1.1 &              51.1$\pm$3.0 &  \underline{42.1}$\pm$2.7 &           54.4$\pm$2.7 &  \textbf{65.3}$\pm$2.6 \\
        & 5 &              44.2$\pm$4.1 &           41.0$\pm$3.7 &              44.0$\pm$2.4 &           40.6$\pm$2.6 &              47.5$\pm$1.7 &  \underline{37.1}$\pm$3.0 &           43.6$\pm$2.6 &  \textbf{51.0}$\pm$2.4 \\
        & 10 &              38.2$\pm$2.0 &           37.9$\pm$2.0 &     \textbf{38.6}$\pm$2.1 &           38.5$\pm$2.7 &  \underline{36.8}$\pm$1.7 &              37.4$\pm$2.4 &           38.3$\pm$2.3 &           37.6$\pm$2.3 \\
\bottomrule
\end{tabular}
}
\vspace{-0.8em}
\end{center}
\end{table*}

\figcifarnone

\section{Experimental results}
\label{sec:experiments}

We compare the class orderings on \mbox{CIFAR-100} considering ten equal tasks trained on \mbox{ResNet-32} from scratch.
Training starts with learning rate (LR) of 0.1, momentum of 0.9, weight decay of \mbox{2e-4}, and a LR scheduler with ten epochs patience and a LR factor of \nicefrac{1}{3} until LR is lower than \mbox{1e-4} or 200 epochs have passed. Method hyperparameter are chosen following the framework from~\cite{de2019continual} for the first 3 tasks, and fixed afterwards.

We compare all orderings proposed in Sec.~\ref{sec:classordering} on Finetuning (FT) and LwF in Fig.~\ref{fig:cifar_none}. Both methods seem to have very little difference in general behaviour on the different orderings. Random, iCaRL seed and minConf provide a better performance after all tasks. The results point to minConf being the most stable, and Random being generally closer to it. After the first task, incTaskConf has the highest performance since it learns the less confusing group of classes. However, after all tasks, it ends up having one of the lowest performances, together with maxConf.

Results on different methods are presented in Tab.~\ref{tab:exemplars-growing}, using 20 exemplars per class with herding selection -- LwF is adapted to use exemplars. As expected, decTaskConf results in the \emph{most confusing} class ordering for the first tasks with the lowest performance. Analogously, incTaskConf achieves the best performance. This is due to not having learned all classes at this point but only the most or least confusing, respectively. This behaviour is different than the one seen in the setting without exemplars, where incTaskConf is only better until task 2. LwF and LUCIR have a similar task 10 overall performance (avg. 28\%), iCaRL follows (avg. 33\%), while BiC and IL2M have a better one (avg. 38\%). In addition, the standard deviation across all orderings is low for BiC and IL2M ({\raise.17ex\hbox{$\scriptstyle\sim$}}2.3). Next comes iCaRL and LUCIR with a bit larger deviation ({\raise.17ex\hbox{$\scriptstyle\sim$}}3.4), and LwF being the least robust ({\raise.17ex\hbox{$\scriptstyle\sim$}}3.6). Interestingly, looking at the best performance at task 10 for each method individually, each of them does well at a different class ordering, thus changing the optic of the result and the ranking of the methods.

\section{Conclusions}
\label{sec:conclusions}

Class orderings for class-IL influence the overall evaluation performance. For a single method the spread between most extreme orderings can be significant. Comparing the orderings, we found that the random ordering obtains among the highest performances when used by non-exemplar methods. The proposed class orderings based on the confusion matrix can be used as a tool for checking robustness of class-IL approaches. A direct extension of this work would be to use different datasets and ordering methods. For a fairer comparison of methods, we recommend to compare methods on several class orderings.

\section*{Acknowledgements}

We would like to thank Xialei Liu for his helpful discussion. Marc Masana acknowledges 2019-FI\_B2-00189 grant from Generalitat de Catalunya.

\bibliography{references}
\bibliographystyle{icml2020}

\newpage
\twocolumn[
\appendixpage
\begin{appendices}
\section{Plots for growing memory setting}
In this appendix we present plots that reflect evaluation of each method from Table~\ref{tab:exemplars-growing} for different class orderings after learning each task.
\\
\\
\end{appendices}
]

\figExpSingleColumn{1.0}{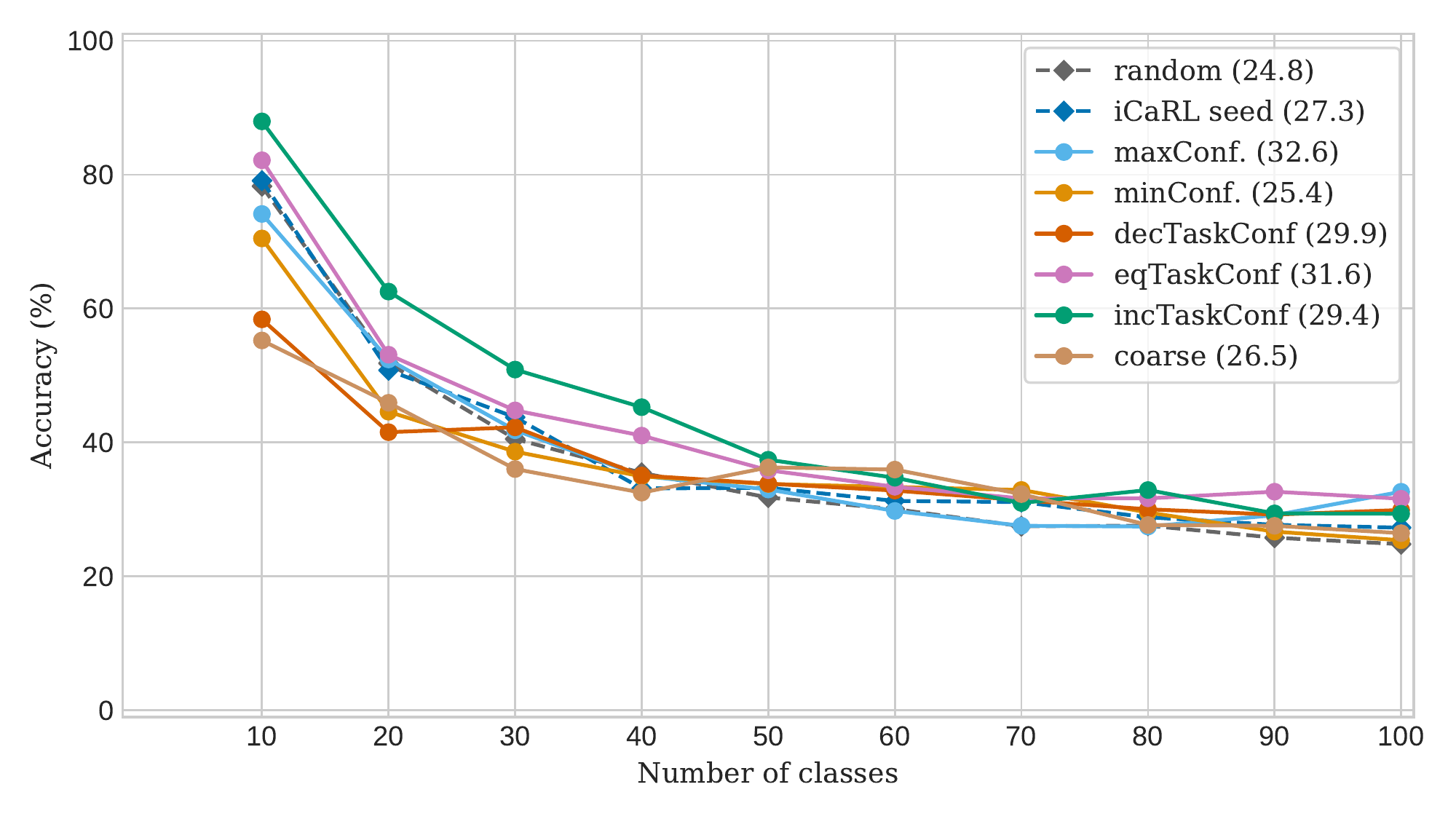}{LwF results for different class orderings for \mbox{CIFAR-100} on \mbox{ResNet-32} from scratch with 20 exemplars per class growing memory.}
\figExpSingleColumn{1.0}{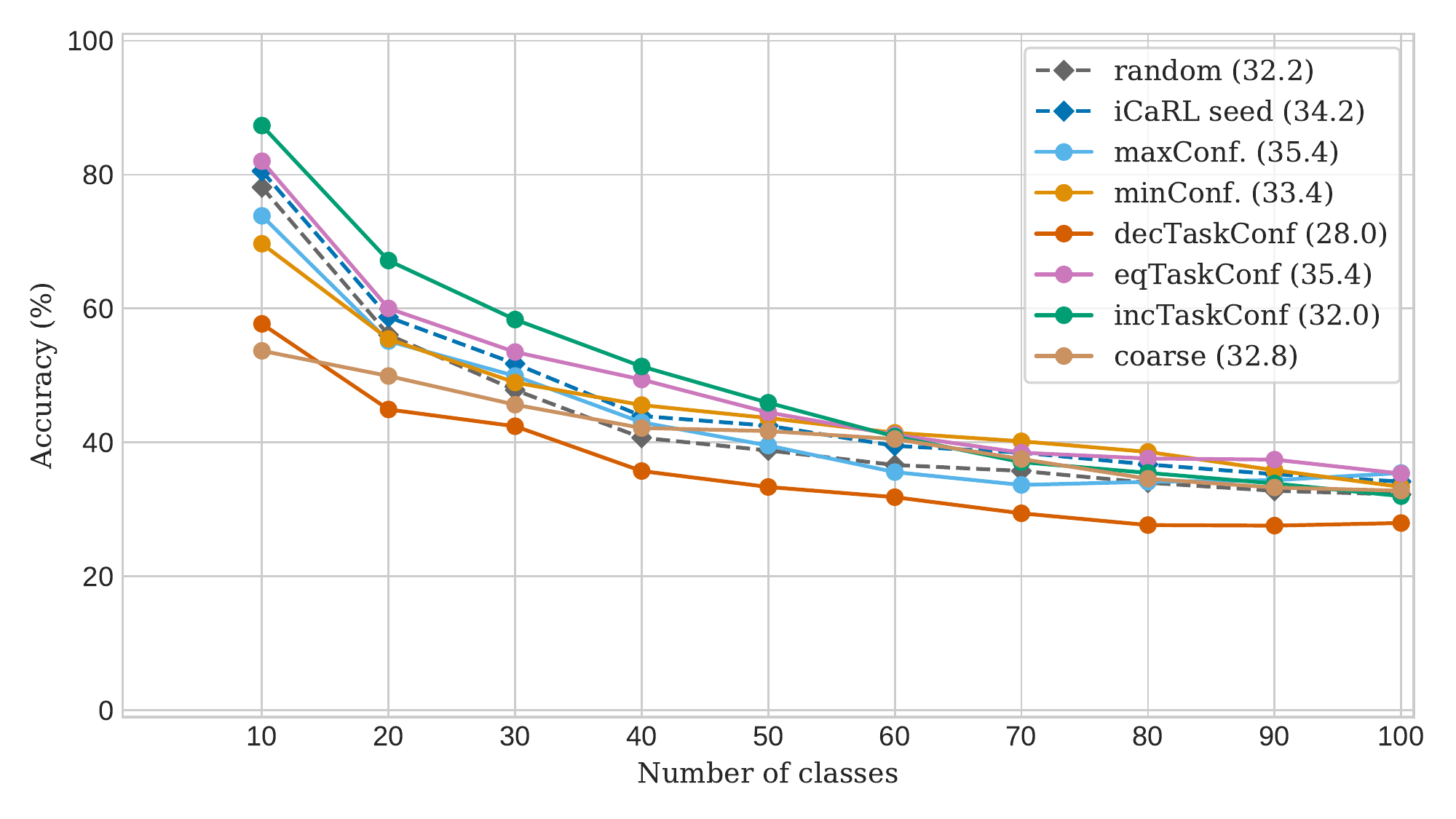}{iCaRL results for different class orderings for \mbox{CIFAR-100} on \mbox{ResNet-32} from scratch with 20 exemplars per class growing memory.}
\figExpSingleColumn{1.0}{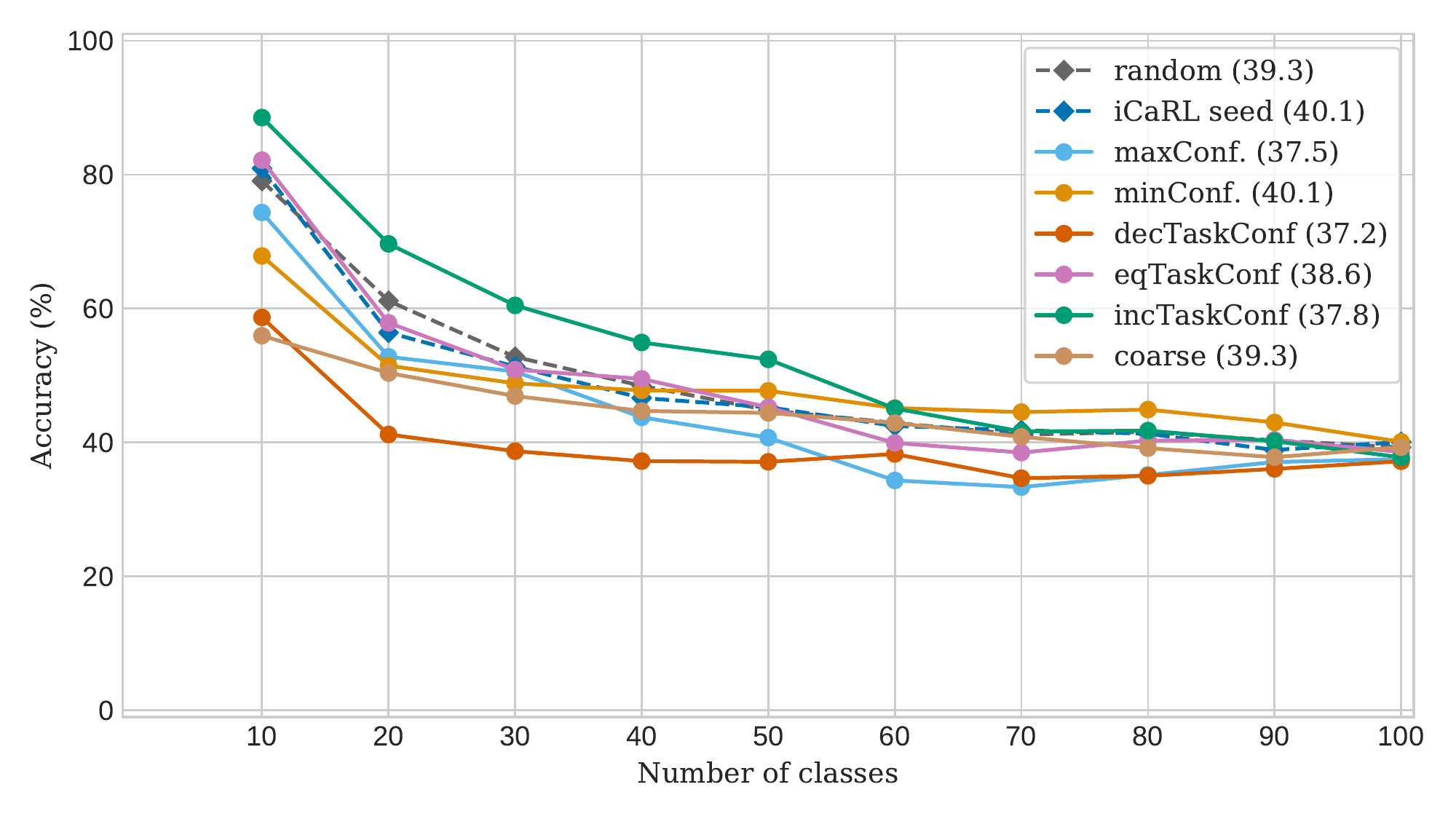}{BiC results for different class orderings for \mbox{CIFAR-100} on \mbox{ResNet-32} from scratch with 20 exemplars per class growing memory.}
\figExpSingleColumn{1.0}{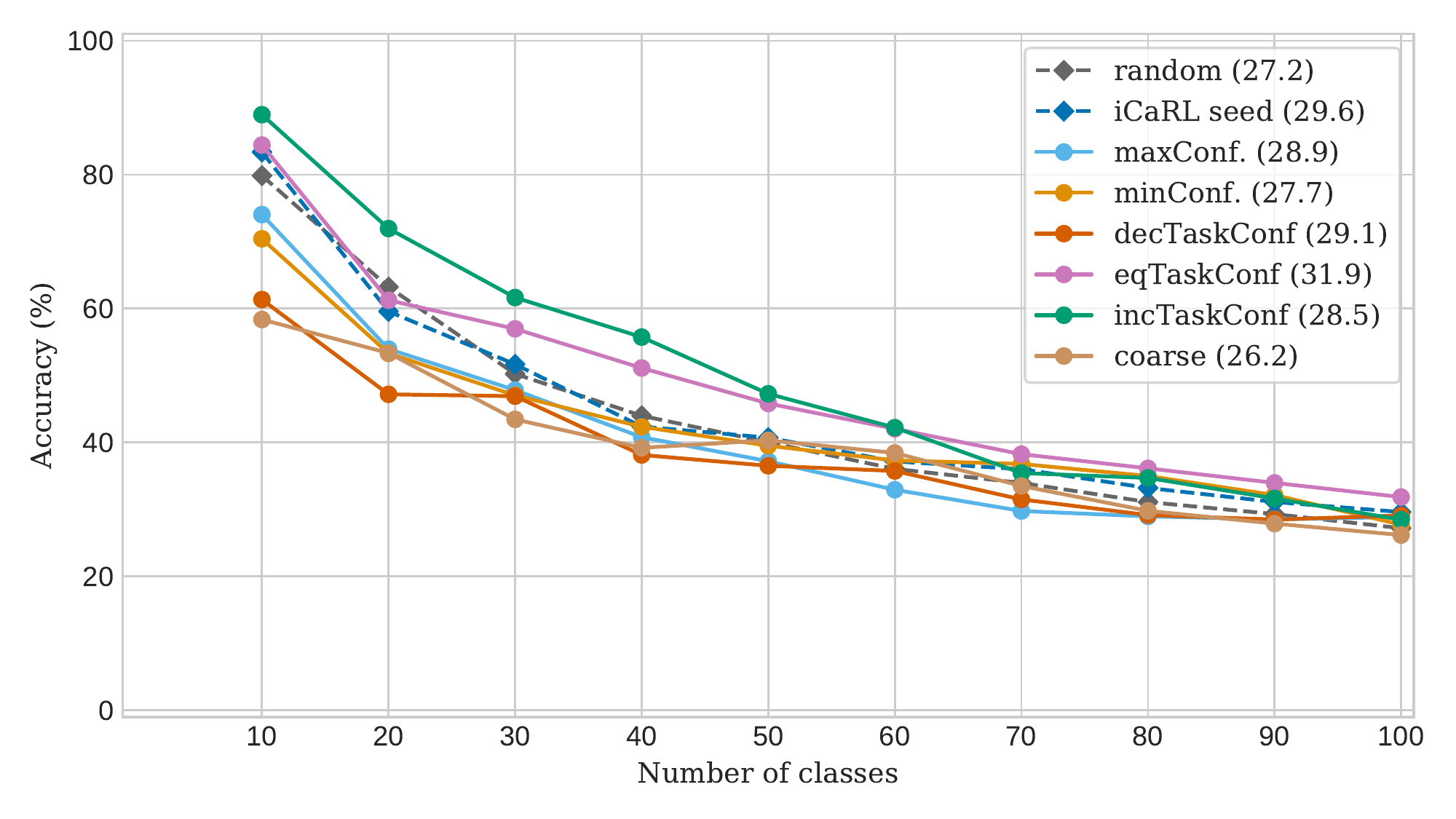}{LUCIR results for different class orderings for \mbox{CIFAR-100} on \mbox{ResNet-32} from scratch with 20 exemplars per class growing memory.}
\figExpSingleColumn{1.0}{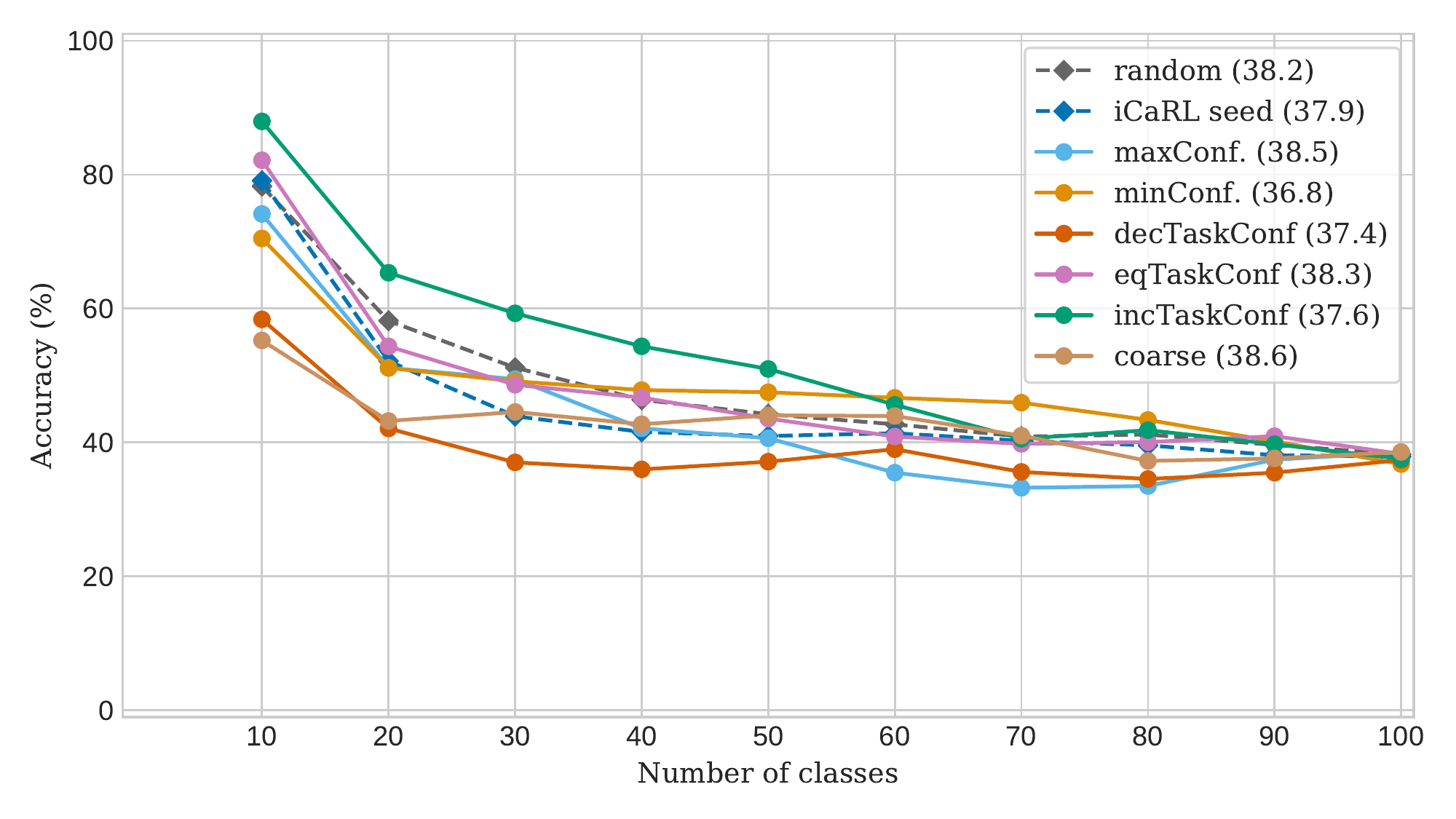}{IL2M results for different class orderings for \mbox{CIFAR-100} on \mbox{ResNet-32} from scratch with 20 exemplars per class growing memory.}

\end{document}